\documentclass[11pt,a4paper]{article}
\usepackage[hyperref]{rocling2024}
\usepackage{times}
\usepackage{latexsym}

\usepackage{microtype}
\usepackage{graphicx}
\usepackage{amsmath}
\usepackage{blindtext}
\usepackage{multirow}

\usepackage[linewidth=1pt]{mdframed}
\definecolor{green}{HTML}{d1f3d3}
\definecolor{red}{HTML}{c20a0a}
\definecolor{orange}{HTML}{FFA500}
\definecolor{lightyellow}{HTML}{FFFACD}

\roclingfinalcopy 
\def\roclingpaperid{20} 

\title{A Cross-Lingual Statutory Article Retrieval Dataset for Taiwan Legal Studies}

\author{
Yen-Hsiang Wang, Feng-Dian Su, Tzu-Yu Yeh, Yao-Chung Fan \\
National Chung Hsing University, Taiwan \\
\texttt{heliart\_sho@hotmail.com, \{sufengdian, vivianyeh12\}@gmail.com, yfan@nchu.edu.tw}
}

\date{}

\begin{document}

\maketitle
\begin{abstract}
This paper introduce a cross-lingual statutory article retrieval (SAR) dataset designed to enhance legal information retrieval in multilingual settings. Our dataset features spoken-language-style legal inquiries in English, paired with corresponding Chinese versions and relevant statutes, covering all Taiwanese civil, criminal, and administrative laws. This dataset aims to improve access to legal information for non-native speakers, particularly for foreign nationals in Taiwan. We propose several LLM-based methods as baselines for evaluating retrieval effectiveness, focusing on mitigating translation errors and improving cross-lingual retrieval performance. Our work provides a valuable resource for developing inclusive legal information retrieval systems.

\end{abstract}


\section{Introduction}

Statutory Article Retrieval (SAR) refers to the task of retrieving relevant statute law articles in response to a legal query. A well-designed SAR system can significantly enhance the efficiency of legal professionals and serve as a cost-effective legal assistant for the public, addressing critical access to justice issues. \citealp{louis2021statutory} highlights a collaboration with legal experts to address limited legal data and introduces the Belgian Statutory Article Retrieval Dataset. Recent advancements in SAR methodologies have been proposed by \citealp{louis2023finding,abdallah2023exploring,nguyen2024enhancing}, focusing on improving retrieval efficiency and accuracy.

\begin{figure}[t]
    \centering
    \includegraphics[width=1.15\linewidth]{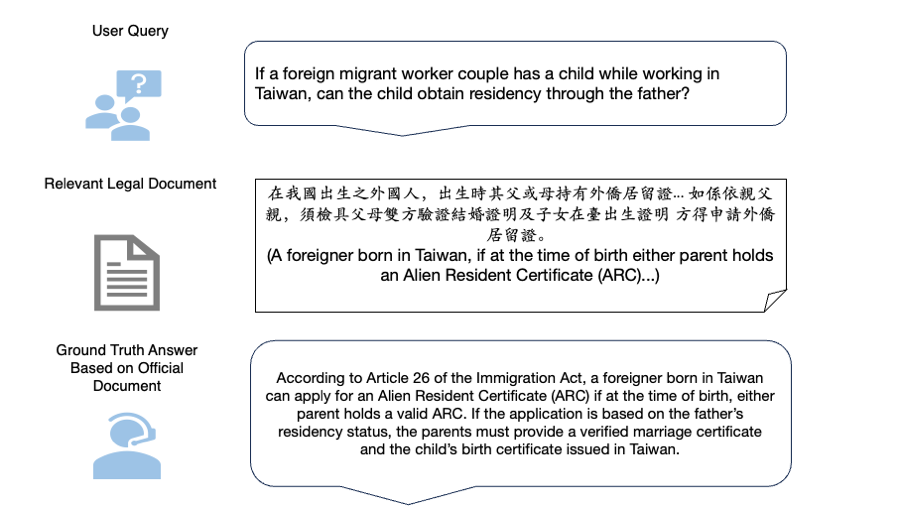}
    \caption{An example in our cross-lingual SAR QA dataset, each instance includes a query and answer in different languages from the data source.}
    \label{fig-data_sample}
\end{figure}

In this paper, we present a cross-lingual statutory article retrieval dataset (called \href{https://github.com/NCHU-NLP-Lab/LawFactsQA-TW}{LawFactsQA-TW}.) designed specifically to address the challenges of legal information retrieval in multilingual contexts. As shown in Figure \ref{fig-data_sample}, our dataset is unique in that it features spoken-language-style legal inquiries, each including an English query, its corresponding Chinese version, the correct relevant statute, and the ground-truth answer for an inquiry. This dataset encompasses all Taiwanese civil law, criminal law, and administrative regulations, providing comprehensive coverage of the legal landscape in Taiwan.

A potential use case for this dataset is for foreign nationals in Taiwan who may wish to inquire about legal rights or obligations in their own language. For instance, a person might ask in English whether their spouse is allowed to work in Taiwan. An ideal retrieval system would then use this English query to locate the corresponding legal provisions in Chinese. This kind of cross-lingual legal information retrieval is essential for improving accessibility to legal information for non-native speakers, ultimately supporting a more inclusive legal system.

The primary goals of this paper are to establish a cross-lingual legal retrieval dataset that supports such use cases and to propose several Large Language Model (LLM)-based methods as baselines for evaluating retrieval effectiveness.

\begin{figure*}[h!]
    \centering
    \includegraphics[width=1.\linewidth]{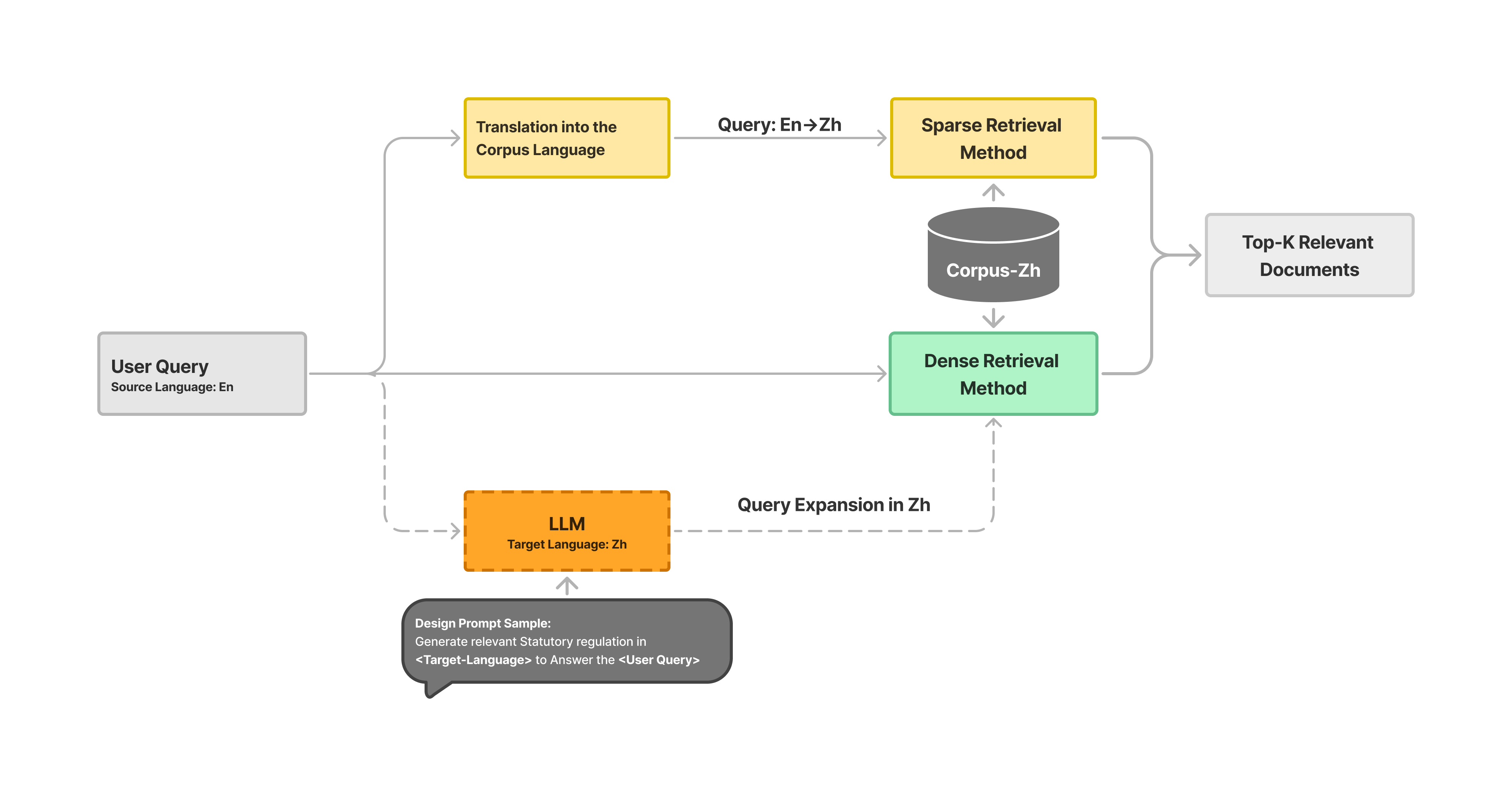}
    \caption{Architecture Diagram of our Cross-lingual Statutory Article Retrieval. The top Branch is \colorbox{lightyellow}{Sparse Retrieval}, which translates the query into the same language within the corpus and then uses a term-based retrieval method. Middle Branch is \colorbox{green}{Dense Retrieval}, which directly uses a multi-lingual embedding model for retrieval. 
    The bottom branch, \colorbox{orange}{LLM-Augmented Retrieval}, leverages large language models for query expansion and dense retrieval methods and then searches the corpus to retrieve the top-K relevant documents.}
    \label{fig-overview-diagram}
\end{figure*}
\begin{figure}[t]
    \centering
    \includegraphics[width=1.\linewidth]{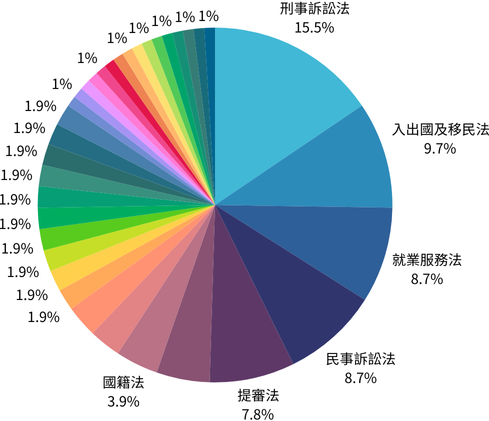}
    \caption{The distribution of relevant laws in our human-labeled dataset, i.e., the Statutory Article corresponding to each question. There are too many items below two percent, so they are not displayed in the chart. For the exact number, please refer to our dataset \href{https://github.com/NCHU-NLP-Lab/LawFactsQA-TW}{LawFactsQA-TW}.}
    \label{fig-human_graph}
\end{figure}

\begin{figure}[t]
    \centering
    \includegraphics[width=1.\linewidth]{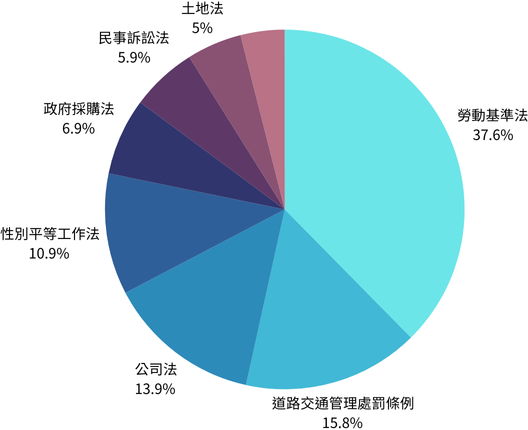}
    \caption{The distribution of relevant laws in our synthetic dataset is relatively balanced because the data generation was based on referencing the search rankings.}
    \label{fig-syn_graph}
\end{figure}

\section{Related Work \label{sec-related_work}}
\subsection{Enhancing Retrieval Performance with LLMs}
Numerous studies have focused on optimizing queries through LLM generation during inference. \cite{ma2023query} propose a framework where an LLM reads retrieval results and rewrites user queries to bridge the gap between the input text and the required knowledge for retrieval. Similarly, \cite{dhuliawala2023chain} introduces a Chain-of-Verification (CoVe) approach, which allows the LLM to decompose the original query into smaller sub-queries, ensuring greater accuracy and reducing hallucinations. 

HyDE \cite{gao2022precise} prompt language models to construct hypothetical documents from a query and then assume it's the relevant document for similarity retrieval.
\cite{chuang2023expand} utilizes a query expansion model to generate a diverse set of queries and then employs a query reranker to select the queries likely to yield improved retrieval results.

In addition to query expansion and rewriting, some studies have investigated using generative LLMs, such as the GPT series, for relevance ranking in Information Retrieval (IR). \cite{sun2023chatgpt} distilled the passage re-ranking capabilities of LLMs into a retrieval model, while \cite{wang2023improving} achieved promising results by obtaining text embeddings from LLMs for retrieval tasks.
The use of LLMs to optimize retrieval, whether through generating rewritten and expanded queries or re-ranking retrieval results, is a highly discussed research direction. This paper's retrieval method design is informed by and builds upon these related studies.

\subsection{LLM for Evaluating Generation Tasks}
Evaluating generation tasks has always been a vital yet challenging topic, and naturally, assessing the generated results is also essential in our retrieval QA task. Recently, several studies have analyzed the strengths and weaknesses of existing LLM-based NLG evaluation methods, such as \cite{gao2024llm, abeysinghe2024challenges, mizrahi2023state}. Given that many recent studies have employed LLMs for evaluation, some research \cite{zheng2024judging, liu2023gpteval} indicates that, when designed with appropriate evaluation methods, LLMs have shown alignment with human judgment.
Thus, we designed an LLM-based evaluation method and validated its alignment with human evaluation to serve as an additional assessment alongside token scores.

\begin{table*}[h!]
\centering
\small
\begin{tabular}{|c|c|c|c|c|c|}
\hline
\textbf{Average Length (Words)} & \textbf{Corpus Passage} & \multicolumn{2}{c|}{\textbf{Question Length}} & \multicolumn{2}{c|}{\textbf{Answer Length}} \\
\cline{3-6}
\textbf{} & \textbf{} & \textbf{Synthetic} & \textbf{Human} & \textbf{Synthetic} & \textbf{Human} \\
\hline
\textbf{zh}& 235 & 26 & 25 & 86 & 180 \\
\hline
\textbf{en}& - & 19 & 20 & - & - \\
\hline
\end{tabular}
\caption{Comparison of Average Corpus Passage Length, Question Length, and Answer Length between Synthetic and Human Data.}
\label{table:corpus_statistics}
\end{table*}

\section{Dataset Construction}

We introduce LawFactsQA-TW\footnote{\href{https://github.com/NCHU-NLP-Lab/LawFactsQA-TW}{LawFactsQA-TW}}, a Frequently Asked Question Answering dataset focusing on Taiwan's laws and regulations. The dataset features bilingual queries (a Traditional Chinese query paired with its corresponding English translation), answers, and relevant legal documents. The knowledge base was constructed by collecting all relevant laws and regulations from the National Regulatory Database\footnote{\href{https://law.moj.gov.tw/api/swagger/ui/index}{National Regulatory Database TW}}.

The dataset comprises a corpus of 5,000 passages from statutory articles, 92 manually annotated QA pairs, and 173 synthetically generated QA pairs. Table \ref{table:corpus_statistics} provides a statistical overview of the dataset, while Figures \ref{fig-human_graph} and \ref{fig-syn_graph} illustrate the distribution of legal regulations in the QA pairs.

The dataset was constructed using the following two methods:

\subsection{Human-Labeled Dataset Collection}
We first collected FAQs from various legal agencies' websites. For queries without a referenced source article, we manually annotated them with the corresponding legal article. In total, 92 such instances were annotated.

\subsection{Synthetic Dataset Generation}
We developed an automated pipeline to generate QA datasets to enhance the experimental data for our study. This pipeline utilizes the \texttt{gpt-4-turbo} model for both question generation and identifying relevant law articles.

The process is divided into two main stages:

\textbf{Stage 1: Data Collection and Question Generation}We obtained search rankings of legal regulations from the Taiwan Ministry of Justice website\footnote{\href{https://law.moj.gov.tw/Hot/Hot.aspx}{Popular legal regulation search logs from the Taiwan Ministry of Justice}}. Using these rankings as scenario contexts, we searched for local news articles related to the most frequently queried statutory articles. We then employed LLMs to generate questions based on the content of these news articles. Each generated question $q$ was paired with a relevant law article $(q, \text{lawArticle})$.

\textbf{Stage 2: Quality Enhancement and Ground Truth Generation}In the second stage, we refined the generated questions by associating them with relevant legal documents and answers. For each question, we presented individual sections of the corresponding law article to the LLM, determining whether an answer could be found within a specific section (creating positive samples). Sections without relevant answers were labeled as negative samples. Questions without a corresponding positive sample were removed from the dataset.

As a result, we compiled a dataset consisting of questions, answers, relevant law sections (positive samples), and negative samples.

\begin{table*}[t]
\centering
\resizebox{\linewidth}{!}{%
\begin{tabular}{lclccc|ccc}
\hline
\textbf{Retrieval Method} & \textbf{Model} & \textbf{Instruction} & \multicolumn{3}{c|}{\textbf{Recall}} & \multicolumn{3}{c}{\textbf{Average Precision}} \\ 
 &  &  & \textbf{@10} & \textbf{@20} & \textbf{@50} & \textbf{@10} & \textbf{@20} & \textbf{@50} \\ \hline
Sparse & X & X & 0.188 & 0.254 & 0.408 & 0.098 & 0.105 & 0.113 \\ \hline
Dense & X & X & 0.348 & 0.446 & 0.626 & 0.207 & 0.217 & 0.230 \\ \hline
\multirow{8}{*}{LLM-Augmentation} & \multirow{2}{*}{GPT-3.5-Turbo} & Answer Expansion & 0.375 & 0.483 & 0.625 & 0.233 & 0.246 & 0.255 \\
 &  & Statutory Article Expansion & 0.423 & 0.554 & 0.687 & 0.258 & 0.274 & 0.284 \\ \cline{2-9} 
 & \multirow{2}{*}{Breeze} & Answer Expansion & 0.359 & 0.458 & 0.636 & 0.203 & 0.217 & 0.230 \\
 &  & Statutory Article Expansion & 0.414 & 0.528 & 0.676 & 0.246 & 0.263 & 0.273 \\ \cline{2-9} 
 & \multirow{2}{*}{GPT-4-0125} & Answer Expansion & 0.393 & 0.510 & 0.612 & 0.206 & 0.225 & 0.232 \\
 &  & Statutory Article Expansion & 0.448 & 0.572 & 0.697 & 0.262 & 0.282 & 0.293 \\ \cline{2-9} 
 & \multirow{2}{*}{Taide} & Answer Expansion & 0.430 & 0.546 & 0.673 & 0.251 & 0.268 & 0.278 \\
 &  & Statutory Article Expansion & 0.455 & \textbf{0.576} & \textbf{0.729} & \textbf{0.302} & \textbf{0.318} & \textbf{0.330} \\ \hline
LLMs Re-ranking & Breeze  & - & \textbf{0.472} & - & - & - & - & - \\ \hline
\end{tabular}%
}
\caption{Performance on Human-Labeled Legal Retrieval Tasks.}
\label{tab-legal-retrieval}
\end{table*}

\begin{table*}[t]
\centering
\resizebox{\linewidth}{!}{%
\begin{tabular}{lccccc|cccc}
\hline
\multirow{2}{*}{\textbf{Retrieval Method}} & \multirow{2}{*}{\textbf{Model}} & \multirow{2}{*}{\textbf{Instruction}} & \multicolumn{3}{c|}{\textbf{Token Score}} & \multicolumn{4}{c}{\textbf{LLM Based Eval}} \\ 
 &  &  & \textbf{BLUE-1} & \textbf{BLUE-2} & \textbf{BLUE-3} & \textbf{Score = 1} & \textbf{Score = 0.5} & \textbf{Score = 0} & \textbf{overall} \\ \hline
Sparse & X & X & 0.140 & 0.044 & 0.018 & 32 & 50 & 10 & 0.620 \\ \hline
Dense & X & X & \textbf{0.165} & 0.053 & 0.024 & 39 & 43 & 10 & 0.658 \\ \hline
\multirow{8}{*}{\begin{tabular}[c]{@{}l@{}}LLM-Augmentation\end{tabular}} & \multirow{2}{*}{GPT-3.5-Turbo} & Answer Expansion & 0.132 & 0.072 & 0.046 & 41 & 46 & 5 & 0.696 \\
 &  & Statutory Article & 0.137 & 0.074 & 0.048 & 34 & 56 & 2 & 0.674 \\ \cline{2-10} 
 & \multirow{2}{*}{GPT-4-0125} & Answer Expansion & 0.126 & 0.065 & 0.040 & 34 & 52 & 7 & 0.652 \\
 &  & Statutory Article Expansion & 0.138 & 0.075 & 0.049 & 38 & 48 & 6 & 0.674 \\ \cline{2-10} 
 & \multirow{2}{*}{Breeze} & Answer Expansion & 0.137 & 0.079 & 0.055 & 31 & 56 & 5 & 0.641 \\
 &  & Statutory Article Expansion & 0.143 & \textbf{0.083} & \textbf{0.057} & 38 & 49 & 5 & 0.679 \\ \cline{2-10} 
 & \multirow{2}{*}{Taide} & Answer Expansion & 0.128 & 0.070 & 0.046 & 31 & 54 & 5 & 0.630 \\
 &  & Statutory Article Expansion & 0.130 & 0.072 & 0.046 & 37 & 49 & 6 & 0.668 \\ \hline
Ground Truth Regulations  & X & X & 0.218 & 0.131 & 0.092 & 57 & 34 & 1 & 0.804 \\ 
No Regulation Provided  & X & X & 0.141 & 0.049 & 0.021 & 26 & 49 & 17 & 0.549 \\ \hline
\end{tabular}%
}
\caption{Legal QA Evaluation with Token Scores and LLM-Based Evaluation.}
\label{tab-legal-qa-eval}
\end{table*}


\begin{table*}[t]
\centering
\resizebox{\linewidth}{!}{%
\begin{tabular}{lclccc|ccc}
\hline
\textbf{Retrieval Method} & \textbf{Model} & \textbf{Instruction} & \multicolumn{3}{c|}{\textbf{Recall}} & \multicolumn{3}{c}{\textbf{Average Precision}} \\ 
 &  &  & \textbf{@10} & \textbf{@20} & \textbf{@50} & \textbf{@10} & \textbf{@20} & \textbf{@50} \\ \hline
BM25 & X & X & 0.328 & 0.409 & 0.530 & 0.209 & 0.219 & 0.227  \\ \hline
BGE-m3 & X & X & 0.469 & 0.583 & 0.737 & 0.250 & 0.273 & 0.287 \\ \hline
\multirow{8}{*}{BGE-m3 + Query Expansion} & \multirow{2}{*}{GPT-3.5-Turbo} & Answer Expansion & 0.476 & 0.615 & 0.785 & 0.286 & 0.314 & 0.331 \\
 &  & Statutory Article Expansion & 0.578 & 0.715 & 0.829 & 0.364 & 0.394 & 0.409 \\ \cline{2-9} 
 & \multirow{2}{*}{GPT-4-0125} & Answer Expansion & 0.524 & 0.668 & 0.785 & 0.324 & 0.350 & 0.3645 \\
 &  & Statutory Article Expansion  & 0.601 & 0.706 & 0.828 & 0.381 & 0.409 & 0.423  \\ \cline{2-9} 
 & \multirow{2}{*}{Breeze} & Answer Expansion & 0.485 & 0.627 & 0.773 & 0.307 & 0.332 & 0.348 \\
 &  & Statutory Article Expansion  & 0.615 & 0.738 & 0.845 & 0.391 & 0.419 & 0.434 \\ \cline{2-9} 
 & \multirow{2}{*}{Taide} &  Answer Expansion & 0.554 & 0.675 & 0.808 & 0.355 & 0.381 & 0.396  \\
 &  & Statutory Article Expansion & 0.598 & 0.736 & 0.833 & 0.375 & 0.405 & 0.418  \\ \hline
 Re-ranking & Breeze & Statutory Article Expansion & \textbf{0.672} & - & - & - & - & -\\ \hline
\end{tabular}%
}
\caption{Performance on synthetic Cross-lingual Laws and Regulations dataset.}
\label{tab-syn-retrieval}
\end{table*}

\begin{figure}[t]
    \centering
    \includegraphics[width=1.\linewidth]{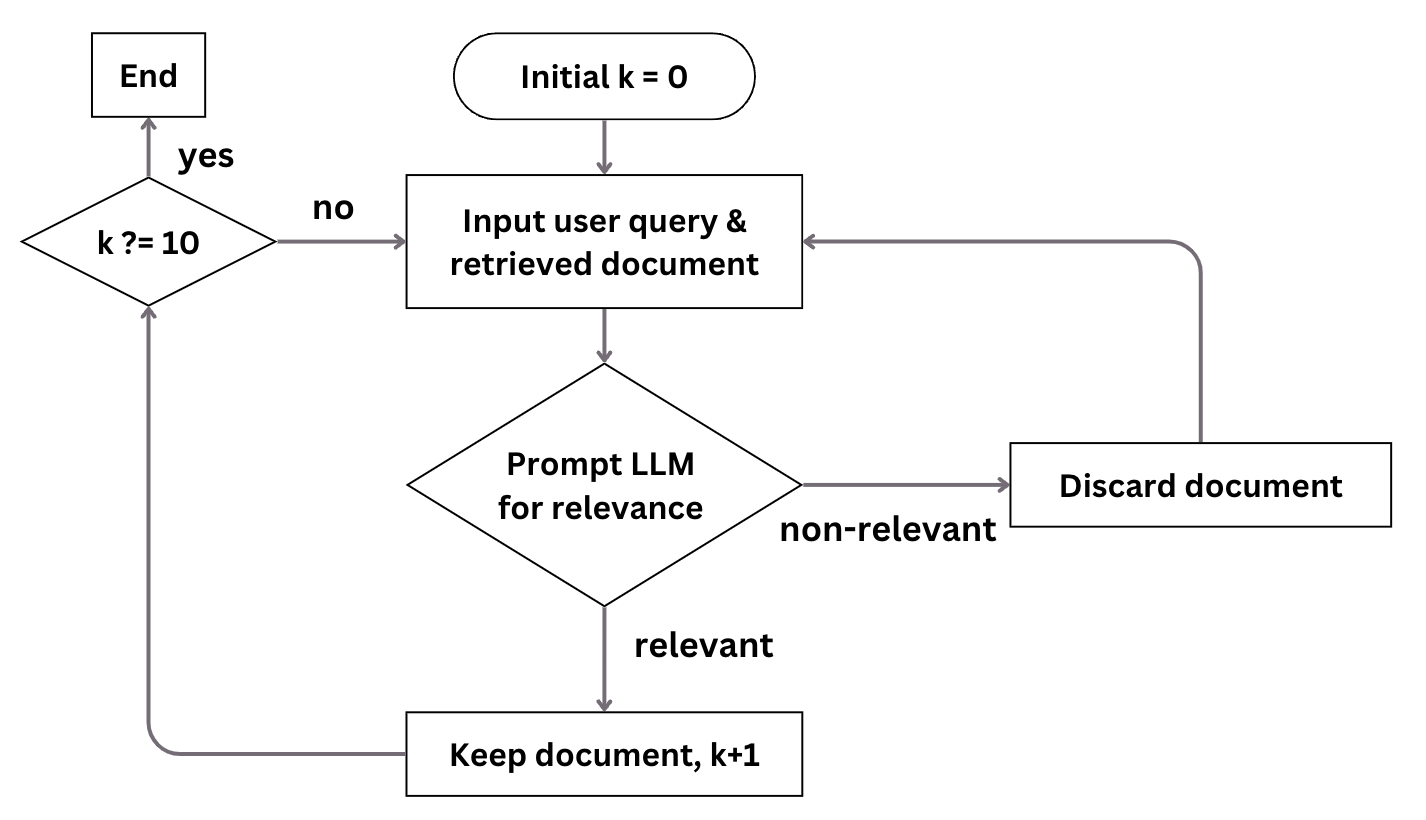}
    \caption{Architecture Diagram for Re-ranking Statutory Article Retrieval Results Based on GPT-4.}
    \label{fig-llm_rerank}
\end{figure}

\section{Methodology}
\label{sec-method}
This section describes three cross-lingual retrieval baselines and the retriever settings used in our experiment. 
Figure \ref{fig-overview-diagram} provides an overview of these three baseline approaches for the cross-lingual SAR task.
The \textbf{Sparse Retrieval} approach involves first translating the query into the same language as the corpus and then performing term-based retrieval to obtain relevant documents. 
In contrast, the \textbf{Dense Retrieval} approach directly uses a multi-lingual embedding model to perform cross-lingual retrieval.

Additionally, we introduce a third method, \textbf{LLM-Augmented Retrieval}, designed to enhance retrieval performance by utilizing LLM-generated content. This method incorporates Answer Expansion, Statutory Article Expansion, and LLM-based Reranking. Please refer to Subsection \ref{sec-LLM-Augmented-Retrieval} for more details.

\paragraph{Retrieval Model}
We use BM25\footnote{Implemented with Pyserini \cite{lin2021pyserini}} as our base model for sparse retrieval. Note that all the queries were translated into the same language as the Statutory Article corpus before performing retrieval. For dense retrieval, we adopt the BGE-m3 model\footnote{\cite{bge-m3}}, a multi-lingual text embedding model, as our embedding-based method for retrieving Statutory Articles.

\subsection{LLM-Augmented Retrieval}\label{sec-LLM-Augmented-Retrieval}
\paragraph{Answer Expansion}
In our first method, we leverage LLMs to directly generate potential answers to the query. Although these generated answers may not always be fully accurate, we use them as hypothetical documents for retrieval. This approach functions similarly to query expansion, where the generated content helps broaden the scope of retrieval, allowing the system to capture documents that are more aligned with the query's intent. By incorporating these generated answers, we aim to improve the accuracy of retrieving relevant statutory articles.

\paragraph{Statutory Article Expansion}
Our second method prompts the LLM to generate statutory article content that could answer the query. Even though the generated content may not be entirely accurate, it can still be used as part of the retrieval process. This helps to enhance the precision of finding the correct statutory articles by aligning the retrieval more closely with the query’s requirements. The generation process supplements the system’s ability to identify relevant legal documents, thus improving retrieval performance.

\paragraph{LLM as Reranker}
In the third method, we use LLMs to rerank the retrieved results. As shown in Figure \ref{fig-llm_rerank}, the LLM re-evaluates the Top-K retrieved statutory articles and replaces irrelevant results with more relevant ones from outside the initial Top-K set. This process is repeated until all retrieved articles are deemed highly relevant to the query. By refining the ranking of results, this method significantly enhances the accuracy of cross-lingual statutory article retrieval.

\section{Experiments and Evaluation Metrics}
In this section, we evaluate the \textbf{LawFactsQA-TW} dataset, which consists of two main tasks: retrieval and question-answering (QA). We utilize \textbf{Recall} and \textbf{Average Precision} as the primary metrics for evaluating the retrieval task, focusing on whether relevant documents are retrieved. While our primary emphasis is on retrieval performance, we also assess the QA task for reference by testing the performance of LLMs. The QA task evaluation includes token-level scoring and other metrics specific to LLM-generated answers.

\subsection{Statutory Article Retrieval Evaluation}
For the retrieval task, we measure performance using \textbf{Recall} and \textbf{Average Precision (AP)} to assess the predicted set of relevant documents, denoted as $D_{\text{predict}} = \{d'_1, \dots, d'_k\}$.

\textbf{Recall} quantifies the ratio of relevant documents retrieved against the total number of relevant documents in the dataset. It provides an indication of how effectively the system retrieves relevant documents, with a higher recall score reflecting better coverage of relevant documents.

\[
\text{Recall} = \frac{|D_{\text{predict}} \cap D_{\text{true}}|}{|D_{\text{true}}|}
\]

\textbf{Average Precision (AP)} offers a more refined evaluation by measuring the precision at various recall levels. \textbf{Precision} is defined as the proportion of relevant documents within the retrieved set. AP is calculated by averaging the precision values across all positions in the ranked list of retrieved documents, with weights based on relevance.

\[
\text{AP} = \frac{1}{|D_{\text{true}}|} \sum_{k=1}^{|D_{\text{predict}}|} P(k) \cdot \text{rel}(k)
\]

Where $P(k)$ is the precision at rank $k$, and $\text{rel}(k)$ is an indicator function returning 1 if the document at rank $k$ is relevant, and 0 otherwise. The term $|D_{\text{true}}|$ represents the total number of relevant documents in the dataset.

We evaluate both metrics at different thresholds: @10, @20, and @50.

\subsection{Question-Answering Evaluation}
For the QA task, we tested two models: \textbf{Breeze 7B}\footnote{\cite{MediaTek-Research2024breeze7b}} and \textbf{GPT-3.5-turbo}\footnote{\cite{brown2020language}} to generate answers. We employed two evaluation strategies: reference-based metrics and LLM-based evaluation.

\textbf{Reference-Based Evaluation}: We used the F1 score and BLEU \cite{papineni2002bleu} to measure the similarity between ground-truth answers and the generated answers.

\textbf{LLM-Based Evaluation}: Answers generated by the models were scored using a three-point scale:
\begin{itemize}
    \item A score of \textbf{1} for responses that matched the correct answer or accurately referenced relevant legal regulations.
    \item A score of \textbf{0} for irrelevant or contradictory responses.
    \item A score of \textbf{0.5} for partially correct responses that were neither fully accurate nor entirely incorrect.
\end{itemize}

Figures \ref{fig-score_prompt}, \ref{fig-1-score_example}, \ref{fig-0-score_example}, and \ref{fig-0.5-score_example} in Appendix \ref{sec-appendix} provide detailed scoring examples.

To evaluate the robustness of the LLM-based method, we defined two scenarios:
\begin{itemize}
    \item \textbf{Best Case}: Ground-truth statutory articles are provided during the retrieval process.
    \item \textbf{Worst Case}: No statutory articles are provided.
\end{itemize}

Table \ref{tab-legal-qa-eval} presents the results of these two scenarios, demonstrating how the availability of statutory articles impacts the model's QA performance.

\section{Results}

\paragraph{Statutory Article Retrieval}
As shown in Table \ref{tab-legal-retrieval}, the Sparse Retrieval method (BM25) serves as our baseline, while the Dense Retrieval method (BGE-M3) demonstrates substantial improvements in both Recall and Average Precision. This result highlights the superior semantic matching capabilities of the dense embedding model, which aligns with our expectations.

In the \textbf{LLM-Augmented Retrieval} experiments, when prompts instruct the model to generate relevant Statutory Articles, we observed a further increase in recall compared to using the model’s embeddings alone. This suggests that LLM-generated content can better align with the retrieval intent.

For our re-ranking method, the retrieval performance at top-10 achieved an 8.4\% improvement in Recall compared to the Dense Retrieval baseline, and a 21\% improvement over Sparse Retrieval. These results demonstrate the effectiveness of the re-ranking method in enhancing cross-lingual retrieval, particularly when working with Traditional Chinese Statutory Articles.

In terms of model performance, both the Breeze 7B \cite{MediaTek-Research2024breeze7b} and Taide\footnote{\href{https://taide.tw/index}{https://taide.tw/index}} models, with 7B parameters, outperformed GPT-3.5 in query expansion. Some metrics even surpassed those of GPT-4, which we attribute to the fine-tuning of Breeze and Taide with Taiwanese Traditional Chinese data. This fine-tuning likely enhances their ability to generate Statutory Articles relevant to the query.

\paragraph{Statutory Article Question Answering}
In the token score evaluation, BLEU-1 scores showed only minor differences between the baseline and the Query Expansion method. However, substantial improvements were observed in BLEU-2 and BLEU-3 scores, with the highest performance achieved using Statutory Articles generated by Breeze. We hypothesize that generating relevant regulations via LLMs can produce answers that more closely match official content.

In the LLM-based evaluation, the retrieval method significantly boosted QA scores compared to the worst-case scenario. The best results were achieved using GPT-3.5, which recorded the highest number of correct answers (scored as 1) and the fewest incorrect answers (scored as 0).

\section{Conclusion}
In this paper, we collected and introduced a Cross-lingual Retrieval Question-Answering dataset for Taiwan legal studies. The dataset includes both human-annotated and synthetic data, which can be used for monolingual or cross-lingual retrieval, as well as legal QA model training and testing. Additionally, we introduce baselines using different retrieval methods paired with various LLMs, intending to serve as a reference for RAG applications in other domains or legal NLP tasks.

\section{Limitation}
We acknowledge that our work has limitations that need to be addressed. For instance, legal professionals have not evaluated our synthetic dataset, which may affect the system's credibility and expertise. Additionally, our dataset is based on common queries from the public. Still, Statutory article retrieval (SAR) often includes retrieval needs specific to legal professionals in their work environment. These issues highlight the need for potential collaboration with experts to refine and improve the system further.

\bibliography{rocling2024}
\bibliographystyle{acl_natbib}
\appendix

\section{Appendix}

\subsection{Human and LLM Evaluation Scoring Benchmark}
\label{sec-appendix}
\begin{figure*}[t!]
\begin{mdframed}
    \parbox{.99\textwidth}{
    \textcolor{red}{Question:}\\
    \{Legal Query\}
    \\
    \textcolor{red}{Gold Answer:} \\
    \{Gound-Thruth answer in our dataset\}
    \\
    \textcolor{red}{Answer by LLM:} \\
    \{System output\}   
    \\
    \\
    \colorbox{green}{Scoring benchmark}\\
    \begin{itemize}
    \item Score of 1 (Positive):
        The model's response is similar to the official answer or references the same legal Regulations.
    \item Score of 0.5 (Neutral):
It is unclear whether the model's response is similar to or inconsistent with the official answer.
    \item Score of 0 (Negative):
The model's response is vague (e.g., 'don't know' or 'can't find') or contrary to the official answer.
    \end{itemize}
   
    }   
\end{mdframed}
\caption{Our prompt for Scoring Legal QA Answer.}
\label{fig-score_prompt}
\end{figure*}

\begin{figure*}[t!]
\begin{mdframed}
    \parbox{.99\textwidth}{
    \colorbox{green}{User Question:}\\
    Do foreigners need to apply for a work permit after obtaining a permanent
residency or naturalizing as citizens of Taiwan in our country?
    \\
    \colorbox{green}{Official Answer:} \\
    Foreign professional talents, specific foreign professional talents, and senior
foreign professional talents, as well as their family members who have
permanent residency in our country, do not need to apply for an individual
work permit according to special laws.
    \\
    \colorbox{green}{LLM Response:} \\
    Yes, foreign nationals who have obtained permanent residency permits or
naturalized to obtain Taiwanese citizenship are exempt from applying for a
work permit when engaging in professional work or seeking employment in
Taiwan.  
    \\
    }   
\end{mdframed}
\caption{Example of an Answer Rated as "1".}
\label{fig-1-score_example}
\end{figure*}

\begin{figure*}[t!]
\begin{mdframed}
    \parbox{.99\textwidth}{
    \colorbox{green}{User Question:}\\
    Can I find foreign teachers to teach me English on tutoring websites at home?
    \\
    \colorbox{green}{Official Answer:} \\
    Currently, the Employment Service Act does not allow individuals to apply to
hire foreign teachers; furthermore, according to Article 43 of the Employment
Service Act: "Foreigners are not allowed to work in the territory of the
Republic of China without the employer's application for permission."
    \\
    \colorbox{green}{LLM Response:} \\
    Yes, you can find foreign teachers to teach you English on tutoring websites
from the comfort of your home. Many tutoring platforms offer online English
lessons taught by qualified native English speakers.
    \\
    }   
\end{mdframed}
\caption{Example of an Answer Rated as "0".}
\label{fig-0-score_example}
\end{figure*}

\begin{figure*}[t!]
\begin{mdframed}
    \parbox{.99\textwidth}{
    \colorbox{green}{User Question:}\\
    What does the term “foreign professionals” refer to?
    \\
    \colorbox{green}{Official Answer:} \\
    Foreign professionals in Taiwan can be hired for specialized or technical
work, managerial positions in approved businesses, and teaching roles at various
educational institutions, sports coaching, and other specialized roles
designated by the Ministry of Education.
    \\
    \colorbox{green}{LLM Response:} \\
    According to the legal Regulations, "foreign professionals" refer to foreigners
engaged in professional work in our country while "foreign-specific
professionals" possess special expertise required by our country. "Foreign
senior professionals" are defined in the Immigration Act as senior
professionals needed by our country.
    \\
    }   
\end{mdframed}
\caption{Example of an Answer Rated as "0.5".}
\label{fig-0.5-score_example}
\end{figure*}



\end{document}